\documentclass[runningheads]{llncs}
\usepackage[T1]{fontenc}
\usepackage{graphicx}
\usepackage{microtype}
\usepackage{array}
\usepackage{amsmath}
\usepackage[table]{xcolor}
\usepackage{colortbl}
\usepackage{amssymb}
\usepackage{amsmath}
\usepackage{booktabs}
\usepackage{multirow}
\usepackage{arydshln}
\usepackage{hyperref}
\usepackage{colortbl}
\usepackage{tabulary}
\usepackage{etoolbox}
\usepackage[capitalize]{cleveref}
\usepackage[square,numbers,sort&compress]{natbib}
\setcitestyle{square, comma, numbers, sort&compress}
\DeclareMathOperator*{\argmin}{arg\,min}

\crefname{section}{Sec.}{Secs.}
\Crefname{section}{Section}{Sections}
\Crefname{table}{Table}{Tables}
\crefname{table}{Tab.}{Tabs.}

\begin{document}
\title{Webcam-based Pupil Diameter Prediction Benefits from Upscaling}

\author{
Vijul Shah\inst{1}\orcidID{0009-0008-5174-0793} \and
Brian B. Moser\inst{1, 2}\orcidID{0000-0002-0290-7904} \and
Ko Watanabe\inst{1, 2}\orcidID{0000-0003-0252-1785} \and
Andreas Dengel\inst{1, 2}\orcidID{0000-0002-6100-8255}
}
\authorrunning{Shah et al.}
\institute{RPTU Kaiserslautern-Landau, Germany \and
German Research Center for Artificial Intelligence (DFKI), Germany\\
\email{\{first.second\}@dfki.de}}

%
\maketitle              
%
\begin{abstract}
Capturing pupil diameter is essential for assessing psychological and physiological states such as stress levels and cognitive load. 
However, the low resolution of images in eye datasets often hampers precise measurement. 
This study evaluates the impact of various upscaling methods, ranging from bicubic interpolation to advanced super-resolution, on pupil diameter predictions. 
We compare several pre-trained methods, including CodeFormer, GFPGAN, Real-ESRGAN, HAT, and SRResNet. 
Our findings suggest that pupil diameter prediction models trained on upscaled datasets are highly sensitive to the selected upscaling method and scale.
Our results demonstrate that upscaling methods consistently enhance the accuracy of pupil diameter prediction models, highlighting the importance of upscaling in pupilometry.
Overall, our work provides valuable insights for selecting upscaling techniques, paving the way for more accurate assessments in psychological and physiological research.

\end{abstract}
%
\section{Introduction}
The widespread adoption of eye-tracking technology in daily life is accelerating, as highlighted by innovations like Apple's camera-based eye tracking~\cite{apple2024accessibility, greinacher2020accuracy}.
As a fortunate side-effect, these technologies enable the analysis of human cognitive states, which are deeply connected to observable features in the eyes~\cite{dembinsky2024eye, dembinsky2024gaze}.
While much of the existing research focuses on blink detection~\cite{hong2024robust} and gaze estimation~\cite{o2023toward, yun2022haze, ankur2024appearance}, which employ biomarker usage~\cite{liu2022noncontact}, infrared reflections~\cite{fathi2015camera}, or image analysis techniques~\cite{hisadome2024rotation}, there is comparatively less emphasis on measuring pupil diameters~\cite{sari2016pupil,caya2022development}. 
Yet, accurately capturing pupil size is critical for assessing various physiological and psychological conditions: Recent research shows that the diameter of the pupil can indicate levels of stress~\cite{pedrotti2014automatic}, focus~\cite{ludtke1998mathematical, van2016pupil}, or cognitive load~\cite{kahneman1966pupil, pfleging2016mental, krejtz2018eye}. 
Moreover, pupil size is linked to the activity of the \textit{locus coeruleus}~\cite{murphy2014pupil, joshi2016relationships}, a crucial brain region for memory management over both short and long terms~\cite{kahneman1966pupil, kucewicz2018pupil}. 
It is also vital in other medical contexts, such as evaluating the pupillary light reflex in patients with brain injuries in intensive care settings~\cite{kotani2021novel}. 
Therefore, precise estimation of pupil diameter is essential for advancing the effectiveness of image-based eye-tracking technologies.

The introduction of the EyeDentify~\cite{shah2024eyedentify} dataset, which offers webcam-based eye images with corresponding pupil diameters, marks a significant advancement in pupilometry research. 
Unlike previous datasets~\cite{ni2018remote,khokhlov2020tiny} that were either not publicly accessible or recorded under highly controlled conditions, EyeDentify provides a diverse array of recordings featuring varying seating positions and distances. 
However, the primary challenge with this dataset is the low quality of the images, which can be attributed to the recording camera quality and the small size of the eyes within the images. 
This necessitates the application of image upscaling techniques to enable the effective use of deep neural networks for pupil diameter prediction.

In this work, we explore the impact of various image Super-Resolution (SR) techniques on the accuracy of webcam-based pupil diameter predictions. 
Image SR aims to transform low-resolution images into high-resolution counterparts, potentially enhancing the clarity and detail of visual data used in training models for more accurate pupil diameter estimation~\cite{moser2023hitchhiker}.
We demonstrate that employing advanced, pre-trained SR models can substantially improve the accuracy of pupil diameter predictions in low-quality, webcam-based images.
Yet, we found that different image SR methods affect pupil diameter estimation differently.
The effectiveness of SR methods varied, with some enhancing the features necessary for precise pupilometry more effectively than others.
Nevertheless, we can conclude that using upscaling methods, in general, improves the performance of pupil diameter prediction models.
Overall, our comparative analysis provides clear guidance on selecting appropriate SR techniques for pupilometry.

\section{Related Work}
In this section, we briefly review the usage of image SR as a pre-processing step for downstream tasks and survey the state-of-the-art of pupil diameter estimation.

\subsection{Super-Resolution as Pre-Processing}
Image SR is the process of transforming a LR image into a HR one, effectively solving an inverse problem~\cite{moser2023hitchhiker}. 
More explicitly, a SR model $M_\theta: \mathbb{R}^{\text{H} \times \text{W} \times \text{C}} \rightarrow \mathbb{R}^{s\cdot\text{H} \times s\cdot\text{W} \times \text{C}}$ is trained to inverse the degradation relationship between a LR image $\mathbf{x} \in \mathbb{R}^{\text{H} \times \text{W} \times \text{C}}$ and the HR image $\mathbf{y} \in \mathbb{R}^{s\cdot\text{H} \times s\cdot\text{W} \times \text{C}}$, where $s$ denotes the scaling factor and the degradation relationship can be described by
\begin{equation}
    \label{equ:JPEG_model}
    \centering
    \mathbf{x}=((\mathbf{y}\otimes \boldsymbol{k})\downarrow_s+\boldsymbol{n})_{JPEG_{q}},
\end{equation} 
where $\boldsymbol{k}$ is a blur kernel, $\boldsymbol{n}$ the additive noise, and \emph{q} the quality factor of a JPEG compression.
In a supervised setting, the training is based on a dataset $\mathbb{D}_{SR} = \{ \left( \mathbf{x}_i, \mathbf{y}_i\right)\}^N_{i=1}$ of LR-HR image pairs of cardinality $N$ and on the overall optimization target
\begin{equation}
    \theta^{*} = \argmin_\theta \mathbb{E}_{(\mathbf{x}_i, \mathbf{y}_i) \in \mathbb{D}_{SR}}\lVert M_\theta (\mathbf{x}_i) - \mathbf{y}_i\rVert^2
\end{equation}
Trained SR models are utilized across a wide array of fields, enhancing everything from medical imaging, where increased image clarity can have critical implications for patient care, to satellite imagery that provides more detailed insights into Earth's geography~\cite{song2022deep,tang2021single}. 
In consumer electronics, such as smartphones and high-definition televisions, SR technologies significantly improve the visual quality, creating more engaging and realistic digital experiences~\cite{zhan2021achieving,shi2016real}.
With the rapid advancements driven by deep learning and cutting-edge generative models, the field of image SR has experienced significant progress~\cite{moser2024diffusion,li2023diffusion,bashir2021comprehensive}.
This work, however, does not seek to develop new image SR methodologies. 
Instead, it leverages SR technology as a preprocessing step to enhance the precision of pupil diameter measurements. 

Similar applications of pre-trained SR models for downstream tasks inspire our goal in related fields, such as image recognition~\cite{kim2024beyond,he2024connecting}, remote sensing~\cite{chen2024super}, dataset distillation~\cite{moser2024latent}, and others~\cite{liu2024improving,jiang2024plantsr}. 
For instance, \textit{Chen et al.} utilized image SR to improve the quality of semantic segmentation~\cite{chen2023semantic}. 
In a different context, \textit{Mustafa et al.} adopted image SR as a defensive strategy against adversarial attacks on image classification systems~\cite{mustafa2019image}. 
Similarly, \textit{Na et al.} applied image SR to boost the performance of object classification algorithms~\cite{na2020object}.
By integrating image SR into our workflow, we aim to refine the input data quality, thus enabling more accurate and reliable analyses in pupil diameter estimation.

\subsection{Pupil Diameter Estimation}
\textit{Ni et al.} introduced a method named BINOMAP for estimating pupil diameter, utilizing dual cameras - referred to as master and slave - as a binocular geometric constraint for analyzing gaze images~\cite{ni2018remote}. 
This model is built on Zhang's algorithm, which recorded a mean absolute error of $0.022\pm0.017$mm~\cite{zhang1999flexible}. 
Similarly, \textit{Caya et al.} used a camera positioned 10cm away from the subject’s face to capture facial images. 
These images were then processed on a Raspberry Pi, which involved converting RGB images to grayscale, adjusting contrast and brightness, reshaping images, and applying the Tiny-YOLO algorithm for pupil diameter estimation~\cite{khokhlov2020tiny}.
Their approach resulted in measurement accuracies with a percent difference of 0.58\% for the left eye and 0.48\% for the right eye.
Both works face significant constraints related to specific conditions, including the necessity for dual cameras and maintaining a constant, fixed distance between the face and the camera.
Another major limitation of these works is that their datasets are not publicly available, contrary to the EyeDentify dataset~\cite{shah2024eyedentify}.

\begin{figure*}[t!]
  \centering
  \includegraphics[width=\linewidth]{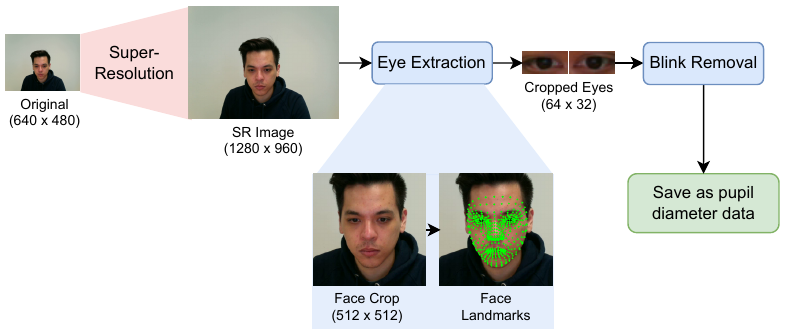}
   \caption{\label{fig:main}Pipeline of our data preprocessing with image SR. As a first step, we super-resolve the raw data with a pre-defined scaling factor (here $2\times$). Next, we used Mediapipe to extract the respective cropped eye images ($64\times32$), left and right, for face detection and landmark localization. Subsequently, we applied blink detection on the cropped eyes using the Eye Aspect Ratio (EAR) and a pre-trained vision transformer for blink detection, as described in EyeDentify \cite{shah2024eyedentify}. Cropped eye images are then saved based on the EAR threshold and model confidence score.}
\end{figure*}

\section{Methodology}


The goal of this work is to apply SR models of the form $M_\theta: \mathbb{R}^{\text{H} \times \text{W} \times \text{C}} \rightarrow \mathbb{R}^{s\cdot\text{H} \times s\cdot\text{W} \times \text{C}}$ to improve the quality of eye images derived from face webcam images, denoted as $\mathbb{D}_{eyes} \subset \mathbb{D}_{faces}$, which is crucial for accurate pupil diameter estimation and cognitive state analysis.
More formally, we aim at constructing $\mathbb{D}^{M_\theta}_{eyes} = \{ \left( M_\theta \left( \mathbf{\hat{x}}_i \right), \mathbf{y}_i\right)\}^N_{i=1}$, where $\left(\mathbf{\hat{x}}_i, \mathbf{y}_i \right) \in \mathbb{D}_{eyes} \subset \mathbb{R}^{\text{H} \times \text{W} \times \text{C}} \times \mathbb{R}$, $\mathbf{\hat{x}}_i \in \mathbb{R}^{\text{H} \times \text{W} \times \text{C}}$ denotes the webcam images of eyes and $\mathbf{y}_i \in \mathbb{R}$ their respective pupil diameter size.
Due to the sparsity of available training data in this eye-monitoring domain~\cite{shah2024eyedentify}, we primarily refer to pre-trained SR models with given parameters $\theta$ instead of training a model $M_\theta$ from scratch.
\autoref{fig:main} illustrates the overall pipeline, which integrates SR, i.e., $M_\theta$, before any face detection, eye localization, cropping, and blink detection.
This revised methodology leverages the strengths of existing SR models while tailoring their application to meet the specific demands of eye feature analysis.

\begin{figure*}[t!]
  \centering
  \includegraphics[width=\linewidth]{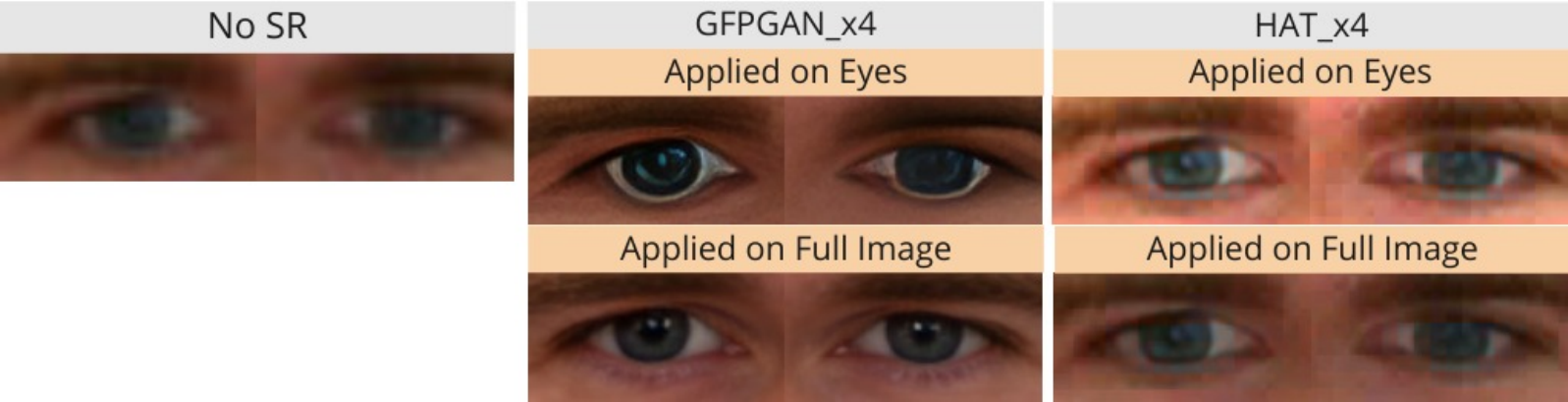}
  \caption{Comparison of applying image SR models on the cropped eye images versus applying them on the entire image. While the SR approximations on the entire image lead to results plausible to the respective input, the SR models applied to the cropped eye images lead to very distinct images. For instance, GFPGAN (left) produces unnatural pupils, whereas HAT (right) emits brightness shifts. 
  }
   \label{fig:ds_imgs_compare}
\end{figure*}

Initially, we planned to apply pre-trained SR techniques directly to isolated images of the left and right eyes, as suggested by the authors of EyeDentify~\cite{shah2024eyedentify}. 
However, this approach faces significant limitations, such as the rarity of eye images in image SR training datasets, e.g., DIV2K~\cite{agustsson2017ntire} or Flicker2K~\cite{timofte2017ntire}.
State-of-the-art SR models like HAT~\cite{chen2023activating} or face SR models like GFPGAN~\cite{wang2021towards} are primarily optimized for everyday or full-face images.
When these models are applied directly to eye images, their effectiveness diminishes due to a mismatch in the data distribution and latent space, which are tailored for the complexities of everyday or entire face features, as shown in \autoref{fig:ds_imgs_compare}.
To address this issue, we propose a more general approach: instead of applying SR directly to eye webcam images $\mathbb{D}^{M_\theta}_{eyes} \subset \mathbb{D}^{M_\theta}_{faces}$, we utilize the entire face webcam images $\mathbb{D}^{M_\theta}_{faces}$. 
Thus, our revised goal is to derive
\begin{equation}
    \mathbb{D}^{M_\theta}_{faces} = \{ \left( M_\theta \left( \mathbf{x}_i \right), \mathbf{y}_i\right)\}^N_{i=1},
\end{equation}
where $\mathbf{x}_i \in \mathbb{D}_{faces} \subset \mathbb{R}^{\text{H} \times \text{W} \times \text{C}}$ denotes the webcam full-face images before any eye-cropping $g:\mathbb{R}^{\text{H} \times \text{W} \times \text{C}} \rightarrow \mathbb{R}^{\text{H}' \times \text{W}' \times \text{C}}$ with $\text{H}' \ll \text{H}$ and $\text{W}' \ll \text{W}$ happened, i.e., $\mathbf{\hat{x}}_i = g \left( \mathbf{x}_i\right)$.
This allows the SR models trained on classical SR datasets $\mathbb{D}_{SR}$ to operate within their optimal data distribution context, i.e., 
\begin{align*}
    &\lVert\mu_{\mathbb{D}_{SR}} - \mu_{\mathbb{D}_{faces}}\rVert^2 \ll \lVert\mu_{\mathbb{D}_{SR}} - \mu_{\mathbb{D}_{eyes}}\rVert^2 \text{ and }\\
    &Tr \left(\Sigma_{\mathbb{D}_{SR}} + \Sigma_{\mathbb{D}_{faces}} - 2\sqrt{\Sigma_{\mathbb{D}_{SR}}\Sigma_{\mathbb{D}_{faces}}}\right) \\&\gg Tr \left(\Sigma_{\mathbb{D}_{SR}} + \Sigma_{\mathbb{D}_{eyes}} - 2\sqrt{\Sigma_{\mathbb{D}_{SR}}\Sigma_{\mathbb{D}_{eyes}}}\right),
\end{align*}
where $Tr{(\cdot)}$ denotes the trace of a matrix, $\mu_{(\cdot)}$ the means and $\Sigma_{(\cdot)}$ the respective covariances.
After enhancing the overall facial images, we proceed with localized feature extraction focused on the eyes. 
This includes precise eye localization, cropping, and subsequent analyses such as blink detection, which we can describe as a function $\varphi_{blink}$ such that $\lvert \mathbb{D}^{M_\theta}_{eyes} \rvert \gg \lvert \varphi_{blink}\left(\mathbb{D}^{M_\theta}_{eyes} \right)\rvert$.

\subsection{SR Techniques}
Regarding SR methodologies, we identify two primary factors that fundamentally influence the performance and outcomes of SR models $M_\theta$: 
the architecture of the models and their training objectives to optimize $\theta$~\cite{moser2024diffusion}. 
Based on the latter, SR models can be broadly categorized into two groups: regression-based models, which typically employ a regression loss, and generative SR models, which utilize adversarial loss mechanisms.
These distinctions are crucial as they result in varying SR approximations, which can subsequently impact the accuracy of pupil diameter estimations. 
To encompass the breadth of techniques available and ensure a comprehensive evaluation, we have selected at least two distinct approaches from each category:
\begin{itemize}
    \item \textbf{Regression-based Models:}
    \begin{itemize}
        \item \textbf{SRResNet:} A general SR method that draws architectural inspiration from ResNet~\cite{he2016deep,ledig2017photo}.
        \item \textbf{HAT:} A state-of-the-art vision transformer designed for image SR~\cite{dosovitskiy2020image,chen2023activating}.
    \end{itemize}
    \item \textbf{Generative Models:}
    \begin{itemize}
        \item \textbf{GFPGAN:} A face-oriented SR GAN model designed specifically to enhance facial features within images~\cite{wang2021gfpgan}.
        \item \textbf{CodeFormer:} A face-oriented VQ-VAE based model~\cite{zhou2022towards}.
        \item \textbf{Real-ESRGAN:} A more generalized SR GAN approach, which is considered to offer robust solutions for generating photorealistic textures and details in everyday situations~\cite{wang2022realesrgan}.
    \end{itemize}
\end{itemize}

\subsection{EyeDentify\texttt{++}}
\begin{figure*}[t!]
  \centering
  \includegraphics[width=\linewidth]{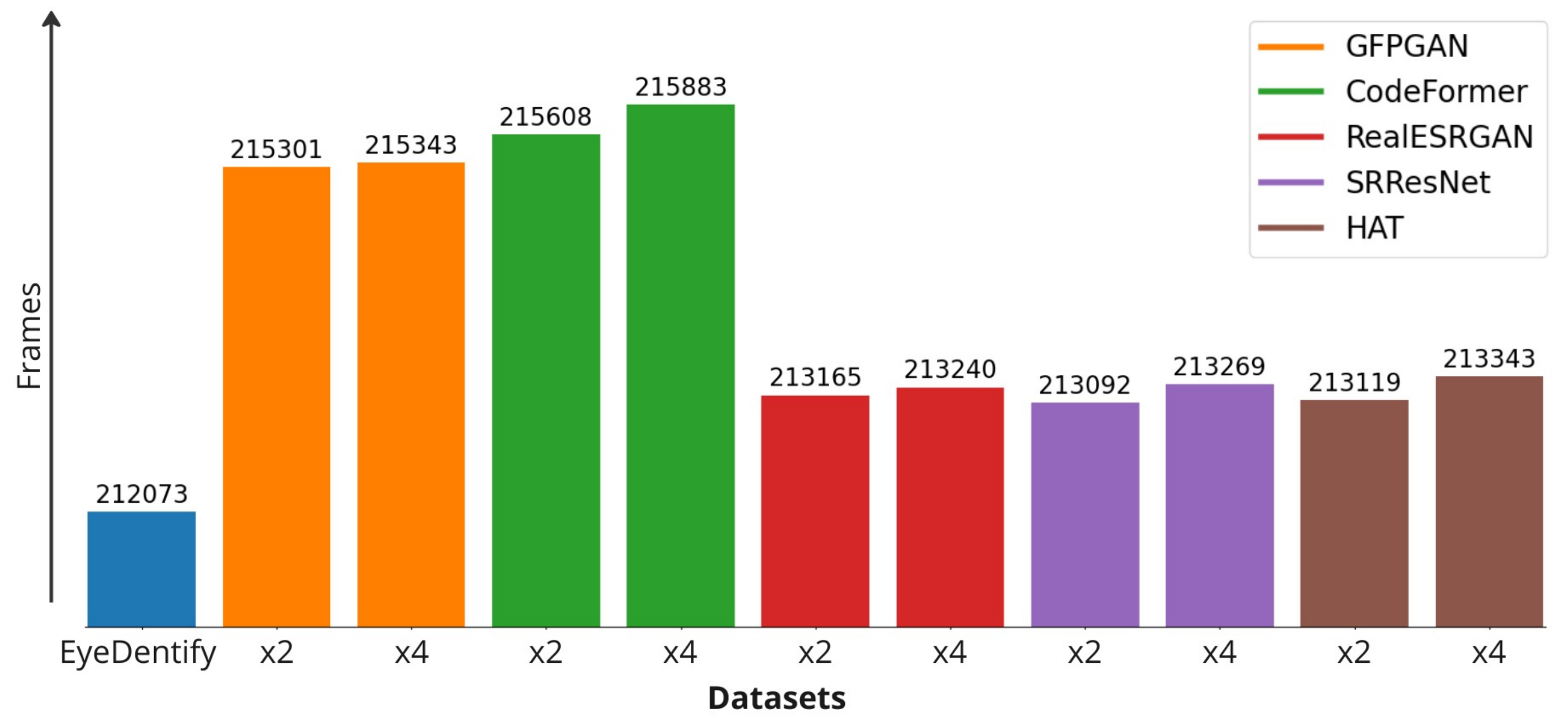}
  \caption{Comparison of applying pre-trained SR models on the EyeDentify Dataset. 
  }
   \label{fig:datasets_counts}
\end{figure*}

As a result of the examination of GFPGAN, CodeFormer, Real-ESRGAN, HAT, and SRResNet SR models for pupil diameter estimation, we can create five additional datasets containing left and right eye images separately, which we call EyeDentify\texttt{++}~\footnote{https://vijulshah.github.io/webcam-based-pupil-diameter-estimation/}. 
Due to the different SR approximations, the later stages, where we recognize faces, crop eyes, and detect blinks, result in retaining and discarding different amounts of images.
More formally, $\lvert \varphi_{blink}\left(\mathbb{D}^{GFPGAN\times2}_{eyes}\right) \rvert \neq \lvert \varphi_{blink}\left(\mathbb{D}^{HAT\times2}_{eyes}\right)\rvert$.
\autoref{fig:datasets_counts} compares the number of images in the original dataset with those in the SR datasets after blink detection. 
The results indicate that S enhances the accuracy of blink classification by improving the calculation of the EAR ratio through clearer eye landmark detection on the 2x and 4x up-scaled images and providing higher-quality images for feature extraction in the subsequent blink detection phase~\cite{shah2024eyedentify}.

\section{Experiments}

In this section, we present our experimental part, which consists of model and training details as well as quantitative and qualitative results.

\subsection{Model Details}
For pupil diameter prediction, we employed the same regression models as suggested in EyeDentify~\cite{shah2024eyedentify}: 
ResNet18, ResNet50, and ResNet152, with the same model configuration and processing steps. 
The datasets created through SR methods were used to train and evaluate these ResNet models. 
We upscaled the eye images by 2x and 4x using bi-cubic interpolation to reach 64 x 32 and 128 x 64 dimensions.
We then refined the images using SR models (e.g., GFPGAN, CodeFormer, Real-ESRGAN, HAT, and SRResNet). 

\subsection{Training Details}
We followed the training setup from the original work \cite{shah2024eyedentify}. 
Using 5-fold cross-validation, we trained ResNet18, ResNet50, and ResNet152 from scratch on all datasets for 50 epochs, with a batch size of 128, separately for left and right eyes. 
We used the AdamW optimizer with default settings, a weight decay of $10^{-2}$, and an initial learning rate of $10^{-4}$, which was reduced by 0.2 every 10 epochs. 


\subsection{Results}
\autoref{tab:results} presents 5-fold cross-validation results for ResNet18, ResNet50, and ResNet152 on SRx2 and SRx4 datasets. 
Compared to the original EyeDentify dataset, we can observe that upscaling greatly benefits pupil diameter prediction.

\textbf{Scale Sensitivity.} \autoref{tab:results} reveals a complex relationship between the scale factor and the performance of SR methods. 
There is no consistent trend of improvement or deterioration as the scale increases from ×2 to ×4 across all methods. 

\textbf{Potential Overfitting.} Certain SR methods exhibit exceptional performance in specific configurations but perform poorly in others. 
For instance, while ResNet152 shows improved results with bicubic interpolation at ×2 scale, it tends to overfit with SR at higher scales. 
This variability could indicate overfitting to particular network architectures, highlighting a need for robustness in classifier selection rather than focusing solely on image enhancement.

\textbf{Best Models.} Across different setups, bicubic upsampling frequently achieves optimal performance for both left and right eyes, particularly notable in the ResNet18 architecture. 
However, advanced SR methods like Real-ESRGAN and SRResNet also consistently demonstrate lower error rates, underscoring their potential effectiveness in specific configurations. 
These findings suggest a balanced approach in selecting SR methods, considering both traditional techniques and advanced models based on specific needs.

\begin{figure*}[t!]
  \centering
  \includegraphics[width=\linewidth]{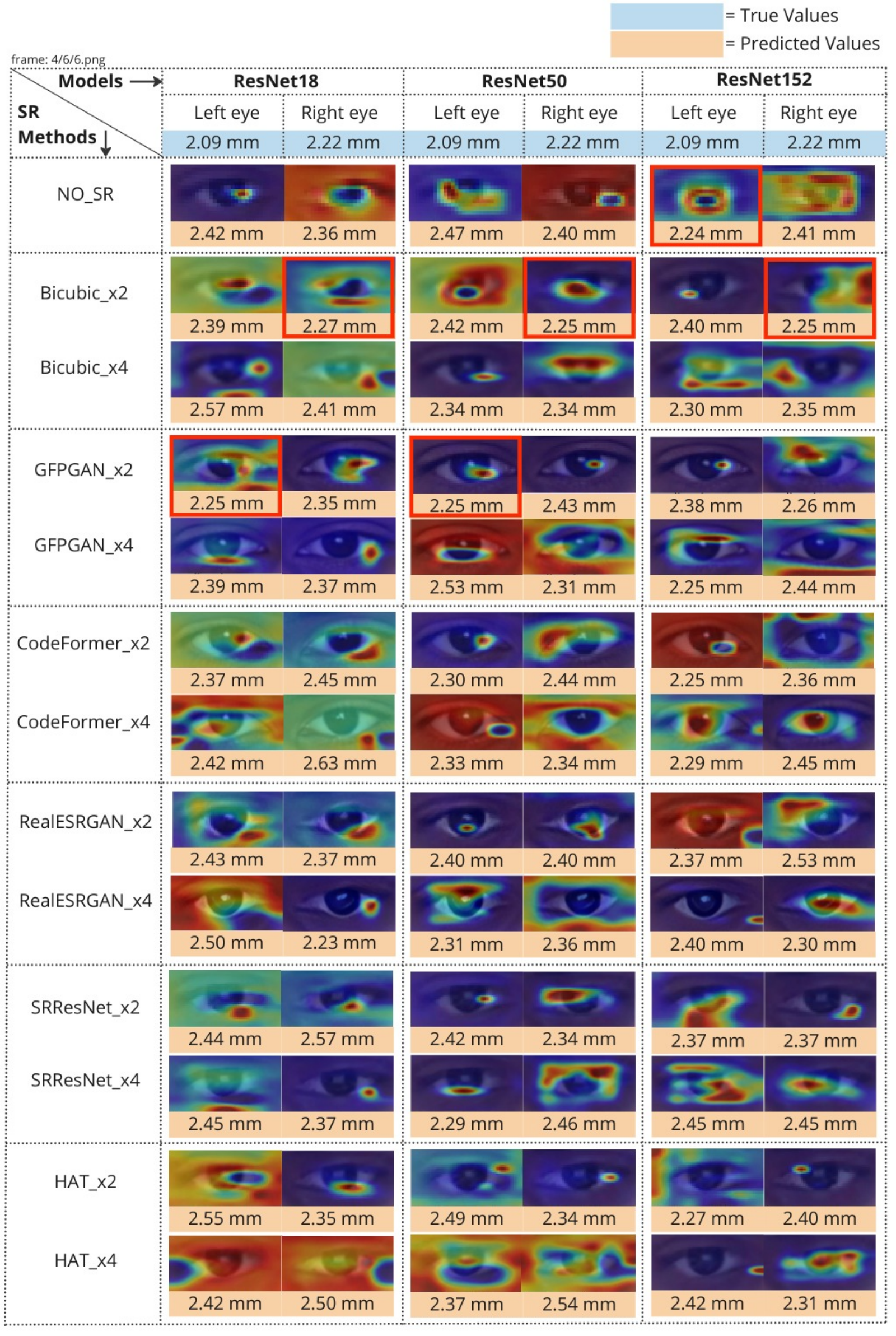}
   \caption{Class Activation Map \cite{zhou2016learning} visualizations for the final convolutional layer of ResNet18, ResNet50, and ResNet152 are shown for a test participant viewing the same display color with No-SR, SRx2, and SRx4 eye images. The true and predicted values represent the original and estimated pupil diameters.}
   \label{fig:cam_viz}
\end{figure*}

 
\textbf{Visualizations.} \autoref{fig:cam_viz} shows the Class Activation Maps (CAM) \cite{zhou2016learning} from the final convolution layer for each model, tested on a participant viewing the same display color across all datasets. The CAM visualizations show that upscaling affects where prediction models focus their attention, with variations in the same image revealing shifts in attention patterns. The top-performing models usually show high activation corresponding to the shape of the eye (see best-performing, boxed examples). Thus, image upscaling influences both the model’s focus and its performance.

\begin{table*}[t!]
\centering
\caption{Quantitative Mean Absolute Error (MAE) $\downarrow$ comparison across different pre-trained SR methods and pupil diameter prediction models for both left and right eyes. The lowest errors are highlighted.}
\renewcommand{\arraystretch}{1.1}
\resizebox{\textwidth}{!}{
\begin{tabular}{|c|c|c|c|c|c|}
\hline
Eye & Scale & Method & ResNet18 & ResNet50 & ResNet152 \\ 
\hline
\multirow{13}{*}{Left} & {$\times$1} & No SR & 0.1329 $\pm$ 0.0235 & 0.1280 $\pm$ 0.0164 & \cellcolor{blue!20}0.1259 $\pm$ 0.0176  \\ \cline{2-6}
& \multirow{6}{*}{$\times$2}& Bi-cubic & 0.1340 $\pm$ 0.0196 & 0.1402 $\pm$ 0.0327 & \cellcolor{blue!30} 0.1225 $\pm$ 0.0166  \\ 
& & GFPGAN & 0.1428 $\pm$ 0.0360 & 0.1486 $\pm$ 0.0195 & \cellcolor{blue!10} 0.1339 $\pm$ 0.0122  \\ 
& & CodeFormer & 0.1328 $\pm$ 0.0245 & 0.1476 $\pm$ 0.0364 & 0.1442 $\pm$ 0.0189  \\ 
& & Real-ESRGAN & \cellcolor{blue!20} 0.1265 $\pm$ 0.0179 & \cellcolor{blue!10} 0.1369 $\pm$ 0.0153 &  0.1384 $\pm$ 0.0195 \\ 
& & SRResNet & \cellcolor{blue!10} 0.1286 $\pm$ 0.0139 & \cellcolor{blue!30} 0.1249 $\pm$ 0.0062 & 0.1391 $\pm$ 0.0261  \\ 
& & HAT & \cellcolor{blue!30} 0.1251 $\pm$ 0.0129 & \cellcolor{blue!20} 0.1277 $\pm$ 0.0241 & 0.1418 $\pm$ 0.0197 \\ \cline{2-6}

& \multirow{6}{*}{$\times$4} & Bi-cubic & \cellcolor{blue!10} 0.1375 $\pm$ 0.0192 & 0.1382 $\pm$ 0.0287 & 0.1497 $\pm$ 0.0275  \\ 
& & GFPGAN & 0.1397 $\pm$ 0.0244 & \cellcolor{blue!30} 0.1230 $\pm$ 0.0122 & \cellcolor{blue!20} 0.1348 $\pm$ 0.0183  \\ 
& & CodeFormer & 0.1383 $\pm$ 0.0170 & 0.1404 $\pm$ 0.0201 & \cellcolor{blue!10} 0.1413 $\pm$ 0.0164  \\ 
& & Real-ESRGAN & \cellcolor{blue!20} 0.1338 $\pm$ 0.0178 & \cellcolor{blue!10} 0.1306 $\pm$ 0.0160 & \cellcolor{blue!30} 0.1316 $\pm$ 0.0183  \\ 
& & SRResNet & 0.1384 $\pm$ 0.0234 & 0.1345 $\pm$ 0.0163 & 0.1509 $\pm$ 0.0242  \\
& & HAT & \cellcolor{blue!30} 0.1330 $\pm$ 0.01191 & \cellcolor{blue!20} 0.1305 $\pm$ 0.0115 & 0.1454 $\pm$ 0.0179 \\ \hline

\multirow{13}{*}{Right} & {$\times$1} & No SR & 0.1548 $\pm$ 0.0273 & 0.1501 $\pm$ 0.0214 & 0.1452 $\pm$ 0.0163  \\ \cline{2-6}
& \multirow{6}{*}{$\times$2} & Bi-cubic & \cellcolor{blue!30} 0.1402 $\pm$ 0.0327 & 0.1558 $\pm$ 0.0214 & 0.1500 $\pm$ 0.0194  \\ 
& & GFPGAN & \cellcolor{blue!20} 0.1470 $\pm$ 0.0328 & 0.1628 $\pm$ 0.0286 & \cellcolor{blue!10} 0.1499 $\pm$ 0.0130  \\ 
& & CodeFormer & 0.1480 $\pm$ 0.0188 & 0.1519 $\pm$ 0.0288 & 0.1542 $\pm$ 0.0423  \\ 
& & Real-ESRGAN & 0.1505 $\pm$ 0.0235 & \cellcolor{blue!10} 0.1502 $\pm$ 0.0154 & 0.1526 $\pm$ 0.0350  \\ 
& & SRResNet & 0.1531 $\pm$ 0.0213 & \cellcolor{blue!20} 0.1490 $\pm$ 0.0328 & \cellcolor{blue!30} 0.1391 $\pm$ 0.0261  \\ 
& & HAT & \cellcolor{blue!10} 0.1477 $\pm$ 0.0321 & \cellcolor{blue!30} 0.1349 $\pm$ 0.0226 & \cellcolor{blue!20} 0.1413 $\pm$ 0.0372 \\ \cline{2-6}

& \multirow{6}{*}{$\times$4} & Bi-cubic & \cellcolor{blue!30} 0.1383 $\pm$ 0.0287 & \cellcolor{blue!30} 0.1319 $\pm$ 0.0222 & \cellcolor{blue!20} 0.1424 $\pm$ 0.0232  \\ 
& & GFPGAN & 0.1595 $\pm$ 0.0157 & 0.1559 $\pm$ 0.0204 & 0.1498 $\pm$ 0.0137  \\ 
& & CodeFormer & \cellcolor{blue!10} 0.1450 $\pm$ 0.0152 & 0.1454 $\pm$ 0.0296 & \cellcolor{blue!10} 0.1441 $\pm$ 0.0211  \\ 
& & Real-ESRGAN & \cellcolor{blue!20} 0.1396 $\pm$ 0.0164 & \cellcolor{blue!20} 0.1321 $\pm$ 0.0375 & 0.1520 $\pm$ 0.0336  \\ 
& & SRResNet & 0.1462 $\pm$ 0.0234 & \cellcolor{blue!10} 0.1345 $\pm$ 0.0163 & 0.1446 $\pm$ 0.0220  \\
& & HAT & 0.1489 $\pm$ 0.0136 & 0.1379 $\pm$ 0.0198 & \cellcolor{blue!30} 0.1369 $\pm$ 0.0236 \\ \hline 
\end{tabular}}
\label{tab:results}
\end{table*}

\section{Limitations}

\begin{figure*}[t!]
  \centering
  \includegraphics[width=\linewidth]{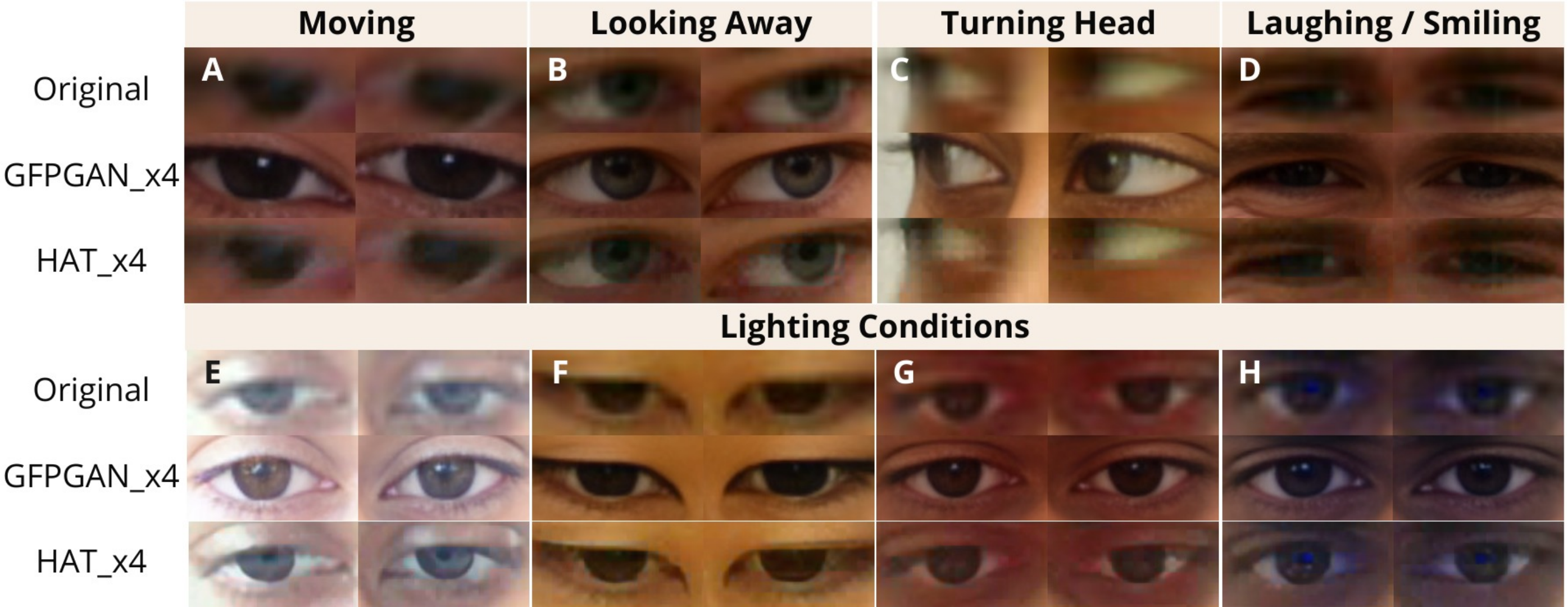}
  \caption{Challenges in estimating pupil diameter without and with SR: Participants A, B, C show head movements and gaze shifts; Participant D shows eye size variation while smiling; Participants E, F, G, H experience different lighting effects—E in bright light, F with a yellow tint, G’s face appearing red, and H’s face appearing blue.
  }
   \label{fig:Eye_Variation_DS}
\end{figure*}

This study faces several challenges, as shown in \autoref{fig:Eye_Variation_DS}. Participants were recorded in natural postures with varying distances from the webcam and no strict positioning guidelines, leading to inconsistencies like movement (A), gaze shifts (B), head/body turns (C), and actions like talking or smiling (D). Differences in eye structure, skin tone, and iris color across diverse nationalities and demographics make it difficult to generalize the model. Variations in lighting and screen color changes further affect the perceived eye and pupil colors (E, F, G, H). Additionally, \autoref{fig:ds_imgs_compare} and \autoref{fig:Eye_Variation_DS} (A, C, E, G, H) highlight that GAN-based models introduce artifacts like glare, altered eye size, and changes in iris color, complicating model training.

\section{Conclusion \& Future Work}
In this work, we investigated the role of SR techniques in enhancing the accuracy of pupil diameter prediction from webcam-based images, which is crucial for assessing psychological and physiological states. 
Our experiments, across multiple upscaling methods and neural network architectures, demonstrate that SR can significantly refine the feature details necessary for more precise pupil measurements.
Key findings indicate that while the benefits of SR are clear, they are not uniformly distributed across different scales and methods. For instance, although traditional bicubic upscaling often performs well, advanced SR techniques like Real-ESRGAN and SRResNet generally provide superior error rates under specific conditions. 
In conclusion, while SR presents a promising avenue for enhancing low-quality, webcam-derived images for pupilometry, it requires nuanced application and thorough validation to fully realize its benefits. 

\section*{Acknowledgements}
This work was supported by the DFG International Call on Artificial Intelligence ``Learning Cyclotron'' (442581111) and the BMBF project SustainML (Grant 101070408).

\bibliographystyle{splncs04}
\bibliography{mybibliography}

\end{document}